\documentclass[conference]{IEEEtran}
\IEEEoverridecommandlockouts
\usepackage{cite}
\usepackage{amsmath,amssymb,amsfonts}
\usepackage{algorithmic}
\usepackage{graphicx}
\usepackage{textcomp}
\usepackage{xcolor}
\usepackage{subcaption}
\def\BibTeX{{\rm B\kern-.05em{\sc i\kern-.025em b}\kern-.08em
    T\kern-.1667em\lower.7ex\hbox{E}\kern-.125emX}}
\usepackage{verbatim}
\begin{document}

\title{Multi-Focus Image Fusion Based on  Spatial Frequency(SF) and Consistency Verification(CV) in  DCT Domain
}

\author{\IEEEauthorblockN{K. S. Krishnendu}
\IEEEauthorblockA{\textit{System Administrator} \\
\textit{catchmysupport.com}\\
Kerala, India \\
krishnendu@catchmysupport.com}
}

\maketitle

\begin{abstract}
Multi-focus is a technique of focusing on different aspects of a particular object or scene. Wireless Visual Sensor Networks (WVSN) use multi-focus image fusion, which combines two or more images to create a more accurate output image that describes the scene better than any individual input image. WVSN has various applications, including video surveillance, monitoring, and tracking. Therefore, a high-level analysis of these networks can benefit Biometrics. This paper introduces an algorithm that utilizes discrete cosine transform (DCT) standards to fuse multi-focus images in WVSNs. The spatial frequency (SF) of the corresponding blocks from the source images determines the fusion criterion. The blocks with higher spatial frequencies make up the DCT presentation of the fused image, and the Consistency Verification (CV) procedure is used to enhance the output image quality. The proposed fusion method was tested on multiple pairs of multi-focus images coded on JPEG standard to evaluate the fusion performance, and the results indicate that it improves the visual quality of the output image and outperforms other DCT-based techniques.  

\end{abstract}


\section{Introduction}
Wireless visual sensor networks (WVSN) have a wide range of applications involving video surveillance, environmental monitoring, and tracking. A high-level analysis of these applications can be utilized effectively in object recognition, particularly in biometrics. Since biometric identifiers are incoherent and difficult to manipulate, a biometric recognition system measures one or more behavioral characteristics or attributes like fingerprints or faces to determine or verify a person’s identity. The accuracy and performance of a biometric system can be affected from the moment of acquisition of the biometric data through its sensor module. A single image with limited depth information cannot solve the problem effectively.  In such cases, Wireless Visual Sensor Networks with multiple sensors can be used to retrieve images with different orientations and illumination of a single object. A centralized fusion center then combines all these source images from multiple sensors to obtain an output image having a more accurate description of that object \cite{nirmala2013multimodal}.

Fingerprints have been utilized for identifying individuals since the early 1900s, with the awareness of the potential individuality of fingerprints due to their unique patterns of ridges and valleys. However, noise can affect the quality of fingerprint images during acquisition. To address this, denoising techniques that explore internal and external correlations have greatly aided automatic fingerprint recognition systems \cite{krishnapriya2017denoising}. Recently, automated face recognition has gained more attention for identifying individuals as faces reveal important attributes such as gender, ethnicity, age, and emotions. However, the main issue with these systems is the dataset used for training images. Implementing image fusion methods to synthesize an output image from available source images can provide a significant impact.

Studies show that face recognition systems have a disparity in accuracy based on race \cite{S_2019_CVPR_Workshops}, gender \cite{albiero2020analysis}, and skin tone \cite{krishnapriya2020issues, krishnapriya2022analysis} even when using balanced datasets \cite{kottakkal2021characterizing}. African American image cohorts have a higher False Match Rate (FMR), while Caucasian images have a higher False Non-Match Rate (FNMR) \cite{kottakkal2021characterizing}. Also, there are differences in impostor and genuine distributions between the image cohorts. For a fixed decision threshold, a higher false match rate of African American and a higher false non-match rate of Caucasian image cohorts reduce the recognition accuracy \cite{S_2019_CVPR_Workshops}. Gender also plays a role, with women having lower recognition accuracy due to lower similarity scores in genuine distribution and higher similarity scores in impostor distribution. Removing facial occlusions improves genuine distribution for women but not impostor distribution \cite{albiero2020analysis}. Finally, skin brightness levels affect matching accuracy, with a mean range resulting in higher accuracy and all other levels resulting in increased false-match and false non-match rates \cite{wu2022face}. So, proper image fusion methods are crucial to address these issues.

Various image fusion methods, such as Discrete Wavelet Transform (DWT) \cite{li1995multisensor}, Shift Invariant Discrete Wavelet Transform (SIDWT) \cite{rockinger1997image}, and Non-Subsampled Contourlet Transform (NSCT) \cite{zhang2009multifocus}, are widely used. However, many of these approaches, which rely on multi-scale transform, are complex and time-consuming. As a result, they may not be suitable for wireless visual sensor networks with limited resources. In such networks, images are typically compressed before being transmitted to other nodes. Using DCT-based standards to save or transmit the source images can significantly reduce computation complexity \cite{haghighat2011multi}. Finally, the fused image is transmitted to an upper node.

Currently, multiple image fusion techniques in the Discrete Cosine Transform (DCT) domain have been developed. Tang et al. \cite{tang2004contrast} proposed two techniques, namely DCT + Average and DCT + Contrast, but they have some drawbacks that result in image quality degradation, such as blurring or blocking artifacts. The algorithm proposed in [8] for DCT + AC - Max \cite{phamila2014discrete} incorrectly selects the right JPEG-coded blocks. Because using the number of higher-valued AC coefficients as a criterion is invalid when most AC coefficients are quantized to zeros during quantization. Another approach \cite{haghighat2010real}, DCT + Variance, considers variance a contrast criterion for fusion, but experimental results in \cite{djamel2006image} indicate that variance performs worse than other focus measures.

This paper proposes a new strategy for image fusion in the DCT domain. This involves using image blocks with higher spatial frequencies to construct the fused image, as spatial frequency reflects the overall active level of an image. A consistency verification process is then applied to enhance the quality of the resulting image. When tested on a dataset of JPEG-encoded image pairs, the results showed a significant improvement in the visual quality of the fused image. This approach is unique in that it uses spatial frequency for the fusion of multi-focus images in the DCT domain rather than the spatial domain. The evaluation metrics indicate that this method outperforms conventional approaches based on DCT and state-of-the-art methods like DWT, SIDWT, and NSCT in terms of visual quality and quantitative parameters. Additionally, this method is easy to implement and computationally efficient, particularly when the source images are coded in JPEG format, making it ideal for wireless visual sensor networks.

The rest of the paper is organized into several sections. Section II will cover the basic concepts behind our proposed algorithm. Section III will outline our approach to image fusion. In Section IV, we will analyze the experimental results. Finally, Section V will provide our conclusions.

\section{DCT Blocks Analysis}
Discrete Cosine Transform (DCT) is one of the most widely used transforms in image compression applications. DCT helps to separate images into parts of differing importance based on the image’s visual quality. It transforms the image from spatial domain to frequency domain. Several standards such as still-JPEG, Motion-JPEG, MPEG, BMP, and TIFF are based on DCT.

Using vector processing, the output matrix of a two-dimensional DCT for an 8 X 8 input matrix is given by: 
\begin{equation} \label{eu_eqn}
F = C.f.C^{t}
\end{equation}

where C is the orthogonal matrix consisting of the cosine coefﬁcients and C\textsuperscript{t}  is the transpose coefﬁcients. The inverse DCT (IDCT) is also defined as:

\begin{equation} \label{eu_eqn}
f = C^{t}.F.C
\end{equation}

Row Frequency (RF) and Column Frequency (CF) of an 8 x 8 image block are given by:
\begin{equation}
RF^{2} = \frac{1}{8*8}\sum_{x=0}^{7}\sum_{y=1}^{7}((f(x, y) - f(x, y - 1))^{2}
\end{equation}

\begin{equation}
CF^{2} = \frac{1} {8*8}\sum_{x=1}^{7}\sum_{y=0}^{7}((f(x, y) - f(x -1, y))^{2}
\end{equation}

The total Spatial Frequency (SF) of an 8 x 8 block in the spatial domain is calculated as:
\begin{equation}
SF^{2} = RF^{2} + CF^{2}
\end{equation}

\section{Proposed Method}
The study of spatial frequency began with the examination of the human visual system. It measures the level of activity in an image and provides an effective contrast criterion for image fusion. Although understanding the human visual system is difficult, the spatial frequency can be calculated easily in the DCT domain. Therefore, we can use spatial frequency as a contrast measure for the source image blocks.

\begin{figure*}[htbp]
\centerline{\includegraphics[width=1.0\textwidth]{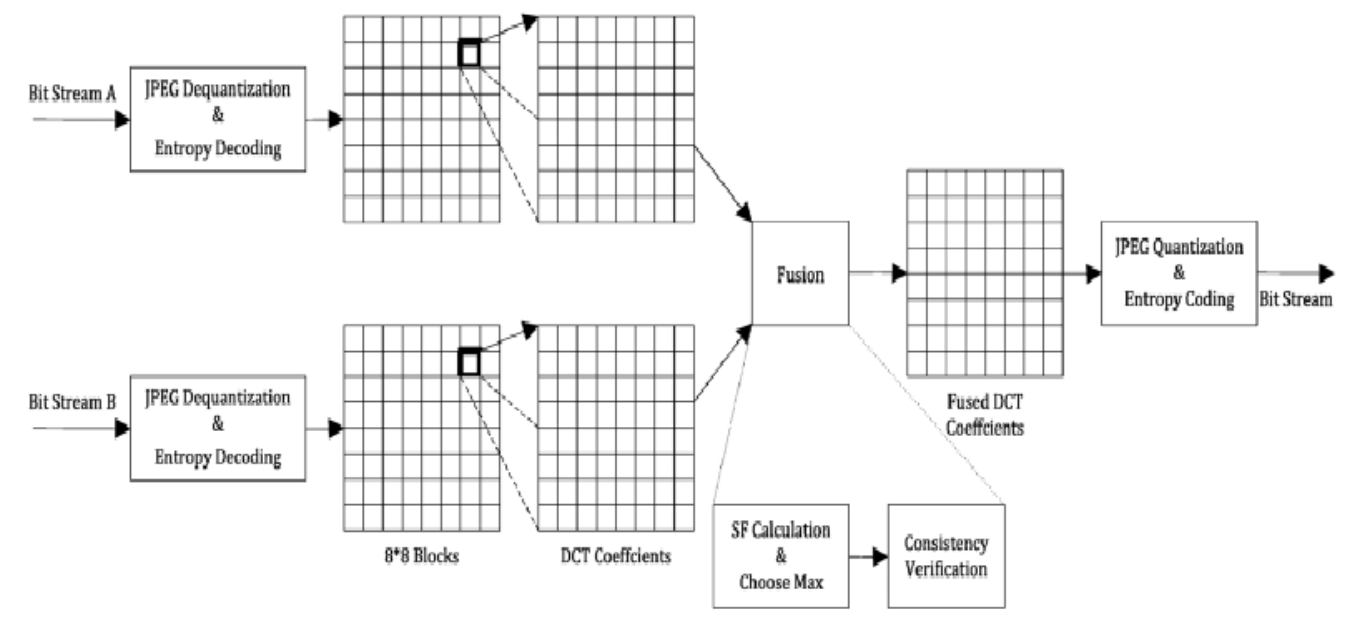}}
\caption{Schematic diagram for fusing images coded in JPEG format}
\label{fig1}
\end{figure*}

Fig.~\ref{fig1} shows the schematic diagram of the proposed multi-focus image fusion technique. For simplicity, it considers only two source images A and B, but the method can be extended for more than two source images. The fusion process consists of the following steps:
\begin{itemize}

    \item Decode and de-quantize the source images, and then divide them into blocks of size 8 x 8. Denote the block pair at location (i, j) by Ai,j and Bi,j respectively.
    \item Compute the spatial frequency of each block and denote the results of Ai,j and Bi,j by SFAi,j and SFBi,j respectively.
    \item Compare the spatial frequencies of two corresponding blocks to decide which should be used to construct the fused image. Create a decision map to record the feature comparison results according to a selection rule:
    \newline
    
    \hspace{1cm}If \emph{Wi,j = 1} Then \emph{SFAi,j $>$ SFBi,j + T}
    
    \hspace{1cm}If \emph{Wi,j = -1} Then \emph{ SFAi,j $>$ SFBi,j + T}
    
    \hspace{1cm}If \emph{Wi,j = 0} Then otherwise\newline

    \hspace{1cm}Here, T is a user-deﬁned threshold. 
    \item Apply a consistency veriﬁcation process to improve the quality of the output image. Use a majority ﬁlter to obtain a reﬁned decision map R:
\begin{equation}
    R_{ij} = \sum_{x=i-1}^{i+1}\sum_{y=j-1}^{j+1}W_{xy}
\end{equation}

    Then, obtain the DCT representation of the fused image based on as:\newline
    
    \hspace{1cm}If \emph{Fi,j = Ai,j} Then \emph{Ri,j $>$ 0}
    
    \hspace{1cm}If \emph{Fi,j = Bi,j} Then \emph{Ri,j $<$ 0}
    
    \hspace{1cm}If \emph{Fi,j = (Ai,j + Bi,j)/2} Then \emph{Ri,j =0}\newline

    \item Quantize the resulting DCT coefﬁcients and then use entropy coding to produce the output bit stream.  

\end{itemize}

\section{Experimental Results and Analysis}
\begin{figure*}[htbp]
\centerline{\includegraphics[width=1.0\textwidth]{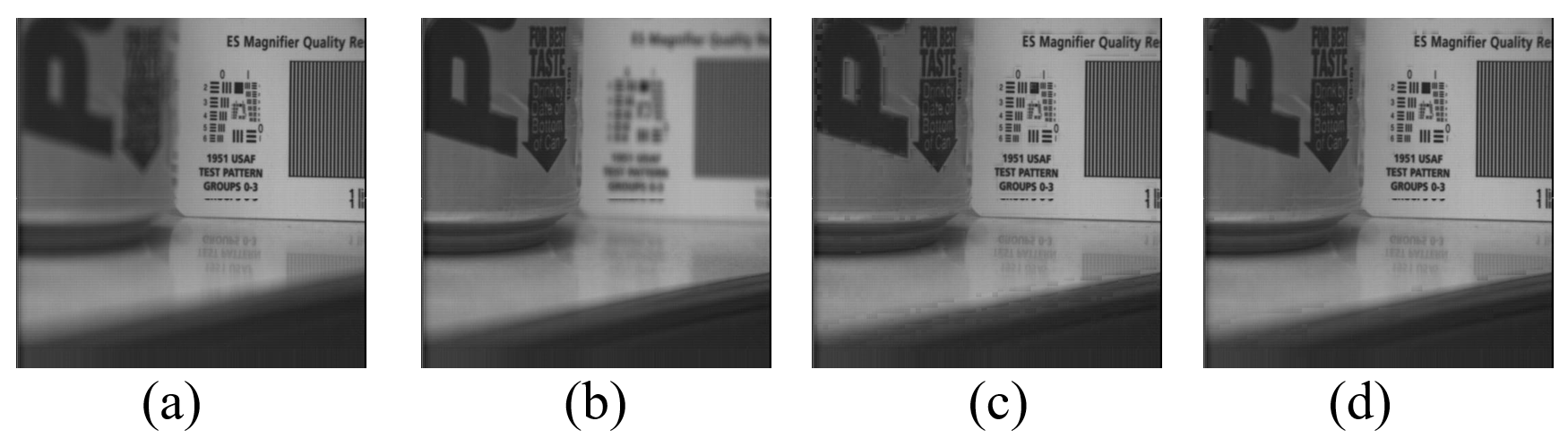}}
\caption{(a) The first source image with focus on the right. (b) The second source image with focus on the left. (c) DCT + SF Result (d) DCT + SF + CV Result}
\label{fig2}
\end{figure*}

This section evaluates the performance of the fusion method by combining various pairs of blurred images. These blurred images are created by filtering standard grayscale images. Complementary regions of the source images are blurred in each pair (See Fig.~\ref{fig2}a and ~\ref{fig2}b). The standard grayscale images serve as the ground truth images.

The measures taken for objective evaluation are Root Mean Square Error (RMSE) and Structural Similarity Measure (SSIM) \cite{wang2004image}. RMSE is the cumulative squared error between the fused image and the referenced image. It should be less for the best output image. It is mathematically calculated as follows:
\begin{equation}
RMSE = \sqrt{\frac{1}{N}\sum_{I=1}^{N}(X_{i} - X_{i})^2}
\end{equation}
where X\textsubscript{i} is the predicted value and N is the total number of observations.

SSIM is used to evaluate salient information that has been transferred into the fused image \cite{wang2004image}. It should be high for the best output image. It is mathematically computed as follows:
\newline
\begin{equation}
SSIM(im1, im2) = \sqrt{\frac{2\mu_{1}\mu_{2} + C_{1}\sigma_{12} + C_{2}}{\mu_{1}^2 + \mu_{2}^2 + C_{1}\sigma_{1}^2 + \sigma_2^2 + C_{2}}}
\end{equation}

where:
\begin{align*}
\mu_1 &= \text{mean value of image 1}\\
\mu_2 &= \text{mean value of image 2}\\
\sigma_1 &= \text{standard deviation of image 1}\\
\sigma_2 &= \text{standard deviation of image 2}\\ 
\text{C1 and C2} &= \text{regularization constants}
\end{align*}

\begin{figure*}[h]
  \centering
  \begin{subfigure}[b]{0.475\linewidth}
      \includegraphics[width=\linewidth]{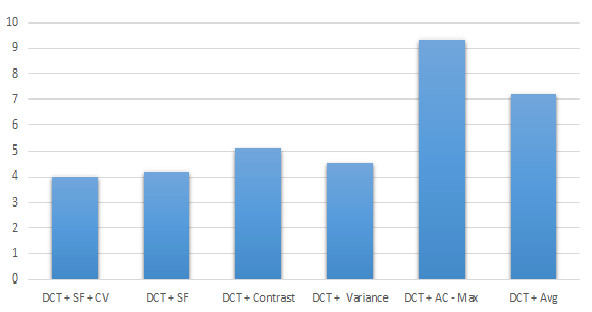}
      \caption{RMSE values}
      \label{fig3}
  \end{subfigure}
  \begin{subfigure}[b]{0.475\linewidth}
      \includegraphics[width=\linewidth]{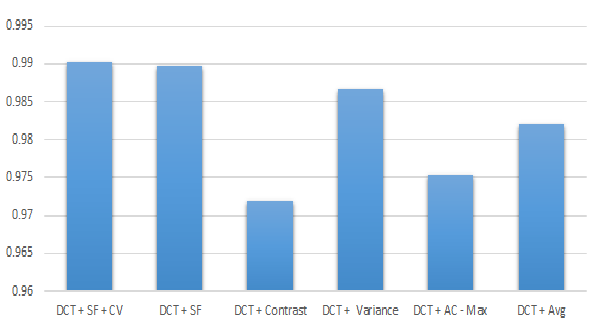}
      \caption{SSIM values}
      \label{fig4}
  \end{subfigure}
  \label{fig_3_4}
  \caption{Comparison of RMSE and SSIM}
\end{figure*}

We compared our image fusion system to existing techniques such as DCT + AC – Max \cite{tang2004contrast}, DCT + Average \cite{haghighat2011multi}, DCT + Contrast \cite{haghighat2011multi}, and DCT + Variance \cite{zhang2009multifocus} to evaluate its performance and feasibility (See Fig.~\ref{fig3} and ~\ref{fig4}). Tables I and II display the average RMSE and SSIM values of 30 experimental images, with the best results highlighted in bold. Our proposed approach, without a CV, outperforms other DCT-based algorithms. Additionally, our CV-enhanced approach produces even better results than all existing image fusion techniques, as demonstrated by its superior RMSE and SSIM values. The following tables TABLE \ref{table:1}, and TABLE \ref{table:2} show the comparison values of RMSE and SSIM, respectively.

\begin{table}[htbp]

\begin{center}
\caption{COMPARISON OF ROOT MEAN SQUARE ERROR (RMSE)}
\label{table:1}
\begin{tabular}{ |c|c|c| } 
 \hline
 \textbf{Method} & \textbf{RMSE of 8 x 8 block} \\
  \hline
 DCT + Avg & 7.213 \\ 
 \hline
 DCT + AC - Max & 9.326 \\ 
 \hline
 DCT + Variance & 4.541 \\ 
 \hline
 DCT + Contrast & 5.103 \\ 
 \hline
 \textbf{DCT + SF} & \textbf{4.220}\\
 \hline
 \textbf{DCT + SF + CV} & \textbf{4.037}
 \\ 
 \hline
\end{tabular}
\end{center}
\end{table}
\begin{table}[htbp]
\begin{center}
\caption{COMPARISON OF STRUCTURAL SIMILARITY MEASURE (SSIM)}
\begin{tabular}{ |c|c|c| }
\label{table:2}

 \textbf{Method} & \textbf{SSIM of 8 x 8 block} \\
  \hline
 DCT + Avg & 0.9821 \\ 
 \hline
 DCT + AC - Max & 0.9753 \\ 
 \hline
 DCT + Variance & 0.9866 \\ 
 \hline
 DCT + Contrast & 0.9719 \\ 
 \hline
 \textbf{DCT + SF} & \textbf{0.9898}\\
 \hline
 \textbf{DCT + SF + CV} & \textbf{0.9902}
 \\ 
 \hline
\end{tabular}
\end{center}
\end{table}

DCT + Average suffers from undesirable blurring effects and the method DCT + Contrast results in blocking artifacts. The method DCT + AC - Max leads to the error selection of the best blocks distinctly. Moreover, DCT + Variance brings about the erroneous selection of some blocks from the blurred image \cite{hossny2010image}.  

From the above experiments and analysis, it is evident that the proposed method is highly effective and surpasses all traditional image fusion approaches in both subjective and objective evaluations. The output fused images have superior visual quality (See Fig.~\ref{fig2}c and ~\ref{fig2}d).

\section{Conclusion}
This paper introduces a new method for merging multi-focus images that use spatial frequency in the DCT domain instead of the spatial domain. The method is evaluated using various metrics, and it has been found that the fusion performance in the DCT domain is better than conventional approaches based on DCT, both in terms of visual quality and quantitative parameters. Additionally, the method is easy to implement and computationally efficient for wireless visual sensor networks.

\bibliographystyle{ieeetr}
{\small
\bibliography{ref}}

% \bibitem{b1} G. Eason, B. Noble, and I. N. Sneddon, ``On certain integrals of Lipschitz-Hankel type involving products of Bessel functions,'' Phil. Trans. Roy. Soc. London, vol. A247, pp. 529--551, April 1955.
% \bibitem{b2} J. Clerk Maxwell, A Treatise on Electricity and Magnetism, 3rd ed., vol. 2. Oxford: Clarendon, 1892, pp.68--73.
% \bibitem{b3} I. S. Jacobs and C. P. Bean, ``Fine particles, thin films and exchange anisotropy,'' in Magnetism, vol. III, G. T. Rado and H. Suhl, Eds. New York: Academic, 1963, pp. 271--350.
% \bibitem{b4} K. Elissa, ``Title of paper if known,'' unpublished.
% \bibitem{b5} R. Nicole, ``Title of paper with only first word capitalized,'' J. Name Stand. Abbrev., in press.
% \bibitem{b6} Y. Yorozu, M. Hirano, K. Oka, and Y. Tagawa, ``Electron spectroscopy studies on magneto-optical media and plastic substrate interface,'' IEEE Transl. J. Magn. Japan, vol. 2, pp. 740--741, August 1987 [Digests 9th Annual Conf. Magnetics Japan, p. 301, 1982].
% \bibitem{b7} M. Young, The Technical Writer's Handbook. Mill Valley, CA: University Science, 1989.
% \end{thebibliography}
% \vspace{12pt}
% \color{red}
% IEEE conference templates contain guidance text for composing and formatting conference papers. Please ensure that all template text is removed from your conference paper prior to submission to the conference. Failure to remove the template text from your paper may result in your paper not being published.

\end{document}